\documentclass[review]{elsarticle}

\usepackage{lineno}
\usepackage{graphicx,color}
\usepackage{amsmath,amssymb,amsfonts}
\usepackage{subcaption}
\usepackage{multirow}
\usepackage{tabularx}
\usepackage{booktabs} 
\usepackage{algorithmic}
\usepackage[ruled,vlined,linesnumbered]{algorithm2e}
\graphicspath{{Figures/}} 
\usepackage{xcolor}
\usepackage{tcolorbox}
\usepackage{comment}
\usepackage{amsmath}
\usepackage{hyperref}
\usepackage{siunitx}
\usepackage{enumitem}
\usepackage{makecell}
\usepackage{multirow,bigstrut}

\modulolinenumbers[0]
\newcommand\Topspace{\rule{0pt}{4ex}}           % Top strut
\newcommand\Bottomspace{\rule[-2ex]{0pt}{0pt}}  % Bottom strut
\newcommand{\blue}[1]{\textcolor{blue}{#1}}

\journal{Journal of Computer Networks}

\begin{document}

\begin{frontmatter}

\title{Aerial Imagery Pile burn detection using Deep Learning: the FLAME dataset}
% \tnotetext[mytitlenote]{Fully documented templates are available in the elsarticle package on \href{http://www.ctan.org/tex-archive/macros/latex/contrib/elsarticle}{CTAN}.}

% \author{Alireza Shamsoshoara}
% \address{Radarweg 29, Amsterdam}
% \fntext[myfootnote]{Since 1880.}

\author[firstaddress]{Alireza Shamsoshoara\corref{mycorrespondingauthor}}
\cortext[mycorrespondingauthor]{Corresponding author}
\ead{Alireza\_Shamsoshoara@nau.edu}

\author[firstaddress]{Fatemeh Afghah}
\author[firstaddress]{Abolfazl Razi}
\author[firstaddress]{Liming Zheng}

\author[secondaddress]{Peter Z Ful{\'e}}
\author[thirdaddress]{Erik Blasch}

\address[firstaddress]{School of Informatics, Computing, and Cyber Systems, Northern Arizona University, Flagstaff, Arizona}
\address[secondaddress]{School of Forestry, Northern Arizona University, Flagstaff, Arizona}
\address[thirdaddress]{Air Force Research Laboratory,  Rome, New York}

\begin{abstract}
Wildfires are one of the costliest and deadliest natural disasters in the US, causing damage to millions of hectares of forest resources and threatening the lives of people and animals. 
Of particular importance are risks to firefighters and operational forces, which highlights the need for leveraging technology to minimize danger to people and property. 
 \textit{FLAME (Fire Luminosity Airborne-based Machine learning Evaluation)} offers a dataset of aerial images of fires along with methods for fire detection and segmentation which can help firefighters and researchers to develop optimal fire management strategies. 

This paper provides a fire image dataset collected by drones during a prescribed burning piled detritus in an Arizona pine forest. The dataset includes video recordings and thermal heatmaps captured by infrared cameras. The captured videos and images are annotated, and labeled frame-wise to help researchers easily apply their fire detection and modeling algorithms.
The paper also highlights solutions to two machine learning problems: (1) Binary classification of video frames based on the presence [and absence] of fire flames. An Artificial Neural Network (ANN) method is developed that achieved a 76\% classification accuracy. (2) Fire detection using segmentation methods to precisely determine fire borders. A deep learning method is designed based on the U-Net up-sampling and down-sampling approach to extract a fire mask from the video frames. Our FLAME method approached a precision of 92\%, and recall of 84\%. Future research will expand the technique for free burning broadcast fire using thermal images.
\end{abstract}


\end{frontmatter}

% \linenumbers

\section{Introduction: Scope, significance, and problem definition}
\label{sec:intro}

Wildfires have caused severe damage to forests, wildlife habitats, farms, residential areas, and ecosystems during the past few years. 
% Several studies including the Australia wildfire, California wildfire, Amazon wildfire show how wildfires can be devastating and more importantly endanger human lives \cite{brando2020amazon, wong2020review, bowman2020wildfires}.
Based on the reports from National Interagency Fire Center (NIFC) in the USA, total number of 51,296 fires burned more than 6,359,641 acres of lands yearly on average from 2010 to 2019 accounting for more than \$6 billion in damages \cite{National39:online, National18:online}. These alarming facts motivate researchers to seek novel solutions for early fire detection and management. In particular, recent advances in aerial monitoring systems can provide first responders and operational forces with more accurate data on fire behaviour for enhanced fire management.

Traditional approaches to detecting and monitoring fires include stationing personnel in lookout towers or using helicopters or fixed-wing aircraft to surveil fires with visual and infrared imaging. Recent research has suggested Internet of Things (IoT) innovations based on wireless sensor networks \cite{toledo2018forest, kaur2019fog, coen2018transforming, afghah2020cooperative, kamhoua2020modeling}, but such networks would require further investment and testing before providing practical information. At broader scales, satellite imagery is widely used for assessing fires globally \cite{huang2020wildfire, friedlingstein2019global}, but typically at relatively coarse resolution and with the availability of repeat images constrained by satellite orbital patterns.

Considering the challenges and issues of these methods, using Unmanned Aerial Vehicles (UAVs) for fire monitoring is gaining more traction in recent years \cite{keshavarz2020real, keshavarz2018towards, keshavarz2019automatic}. UAVs offer new features and convenience including fast deployment, high maneuverability, wider and adjustable viewpoints, and less human intervention \cite{erdelj2017help,Afghah_ACC, aggarwal2020risk,Afghah_INFOCOM, afghah2018reputation}. Recent studies investigated the use of UAVs in disaster relief scenarios and operations such as wildfires and floods, particularly as a temporary solution when terrestrial networks fail due to damaged infrastructures, communication problems, spectrum scarcity, or coalition formation \cite{shamsoshoara2019distributed, shamsoshoara2019solution, shamsoshoara2020autonomous,mousavi2019use,mousavi2018leader}.

Recent advances in artificial intelligence (AI) and machine learning have made image-based modeling and analysis (e.g., classification, real time prediction, and image segmentation) even more successful in different applications \cite{mousavi2017traffic, mousavi2016deep,sarcinelli2019handling,mousavi2016learning}. Also, with the advent of nanotechnology semiconductors, a new generation of Tensor Processing Units (TPUs) and Graphical Processing Units (GPUs) can provide an extraordinary computation capability for data-driven methods \cite{wang2019benchmarking}. Moreover, modern drones and UAVs can be equipped with tiny edge TPU/GPU platforms to perform on-board processing on the fly to facilitate early fire detection before a catastrophic event happens \cite{google_edge_tpu:online, NVIDIAJe4:online}.

Most supervised learning methods rely on large training datasets to train a reasonably accurate model. Studies such as \cite{wu2020transfer} used a fire dataset from public sources to perform fire detection based on pre-trained ANN architectures such as MobileNet and AlexNet. However, that dataset was based on terrestrial images of the fire. To the best of our knowledge, there exists no aerial imaging dataset for fire analysis, something in urgent need to develop fire modeling and analysis tools for aerial monitoring systems. Note that aerial imagery exhibits different properties such as low resolutions, and top-view perspective, substantially different than images taken by ground cameras.

In this paper, we introduce a new dataset as a collection of fire videos and images taken by drones during a prescribed burning slash piles in Northern Arizona.
The images were taken by multiple drones with different points of view, different zoom, and camera types including  regular and thermal cameras. Pile burns can be very helpful to study spot fires and early-stage fires. Pile burns are typically used by forest management for cleaning up forest residues (“slash”) such as branches and foliage from forest thinning and restoration projects. Forest treatments are a key management strategy for reducing fuels and the burning of slash piles is often the most economically efficient and safe means of removing slash. Piles must be monitored by fire managers for a few days after ignition to avoid spread outside the intended burn area. Using automated aerial monitoring systems can substantially reduce the forest management workload.
%\red{Studying and investigating the pile fire can help researchers to have a better understanding for the wildfire prevention. } \fa{not sure if that's an accurate statement. Can you search the literature and provide some references. we dont need to focus on wildfires here. Just need to add a motivation for drone-based prescribed fire monitoring} 

We propose two sample problems to evaluate the use of dataset for real-world fire management problems.
The contributions of this paper include i) proposing the first of its kind aerial imaging dataset for pile burn monitoring which includes both normal and thermal palettes as well as FLAME (Fire Luminosity Airborne-based Machine learning Evaluation), ii) a DL-based algorithm for frame-based fire classification which can be used for early fire detection, and iii) a DL-based image segmentation method for pixel-wise fire masking for fire expansion modeling.
% , and iv) an optimal scheduling approach to observe a set of spotted fires using a minimum number of drones.
The rest of the paper is structured as follows. Section~\ref{sec:dataset} presents the FLAME dataset along with the related information regarding the hardware and data. Section~\ref{sec:method} discusses the methodology based on the two defined challenges, namely fire classification and fire segmentation.
% , and UAV scheduling.
The experiments and results are illustrated in Section~\ref{sec:results} over a variety of metrics. Conclusions and discussion points are provided in Section~\ref{sec:conclusion}.

\section{FLAME Dataset: Hardware and Applicable Data}
\label{sec:dataset}
This section details the hardware used to collect information, the data modalities, and  types of the captured information.
% The location of the prescribed fire is the ...
% \arr{fix this, as Pete said it was not Centennial forest}
% \as{waiting to get new information, it would be great if Dr. Fule has a map of the location, then we can replace it with the previous map as well (Figure~\ref{fig:map}).} 

% Figure~\ref{fig:map} shows the Centennial forest as a green region on the map. This forest is divided in half, one half is located on southwest of Flagstaff town within the Coconino National Forest, the other half which is considered for this study is located on north side of the Flagstaff \cite{NAUsCen31:onlineforest}}. 
%\fa{you can add the info on the vegetation type} \red{Using Centennial forest for the purpose of research requires a permission which is not free. 
Prescribed burning of slash piles is a common occurrence primarily during the winter months in high-elevation forests of the Southwest. Prescribed fires provide excellent opportunities for researchers to collect and update imagery data. The current study shows the results of the first test, and from which is available to continually update the dataset by adding more test results. The test was conducted with fire managers from the Flagstaff (Arizona) Fire Department who carried out a burn of piled slash on city-owned lands in a ponderosa pine forest on Observatory Mesa. The prescribed fire took place on January 16th, 2020 with the temperature of 43$^\circ$F ($\sim$ 6$^\circ$C) and partly cloudy conditions and no wind. 

%*****************************************Figures
% \begin{figure}[bt]
% 	\centering
% 	\includegraphics[width=0.75\columnwidth]{Figures/Map.pdf}
% 	\caption{\red{Northern Arizona University's (NAU) Centennial forest \cite{NAUsCen31:onlineforest}.}\as{waiting for Dr. Fule to update the map if he has any}} 
% % 	\vspace{-10pt}
%     \label{fig:map}
% \end{figure}
% *****************************************Figures

\subsection{Hardware}
\label{subsec:hardware}
This study utilizes different drones and cameras to create a dataset of fire aerial images. Table~\ref{tab:tools_hardware} describes the technical specification of the utilized drones and cameras.

% *****************************************Table
% \renewcommand{\arraystretch}{2.0}
\begin{table*}[bt]
\caption{Technical specification of hardware and tools}
 \centering{
\label{tab:tools_hardware}
\resizebox{0.965\linewidth}{!}{  %fit to windows command
% \color{blue}
% \renewcommand{\arraystretch}{2.5}
\begin{tabular}{c|c}
\toprule
\toprule
\Topspace
\Bottomspace
% \raisebox{-\totalheight}
\multirow{3}{*}[0.2in]{\includegraphics[width=0.2\textwidth]{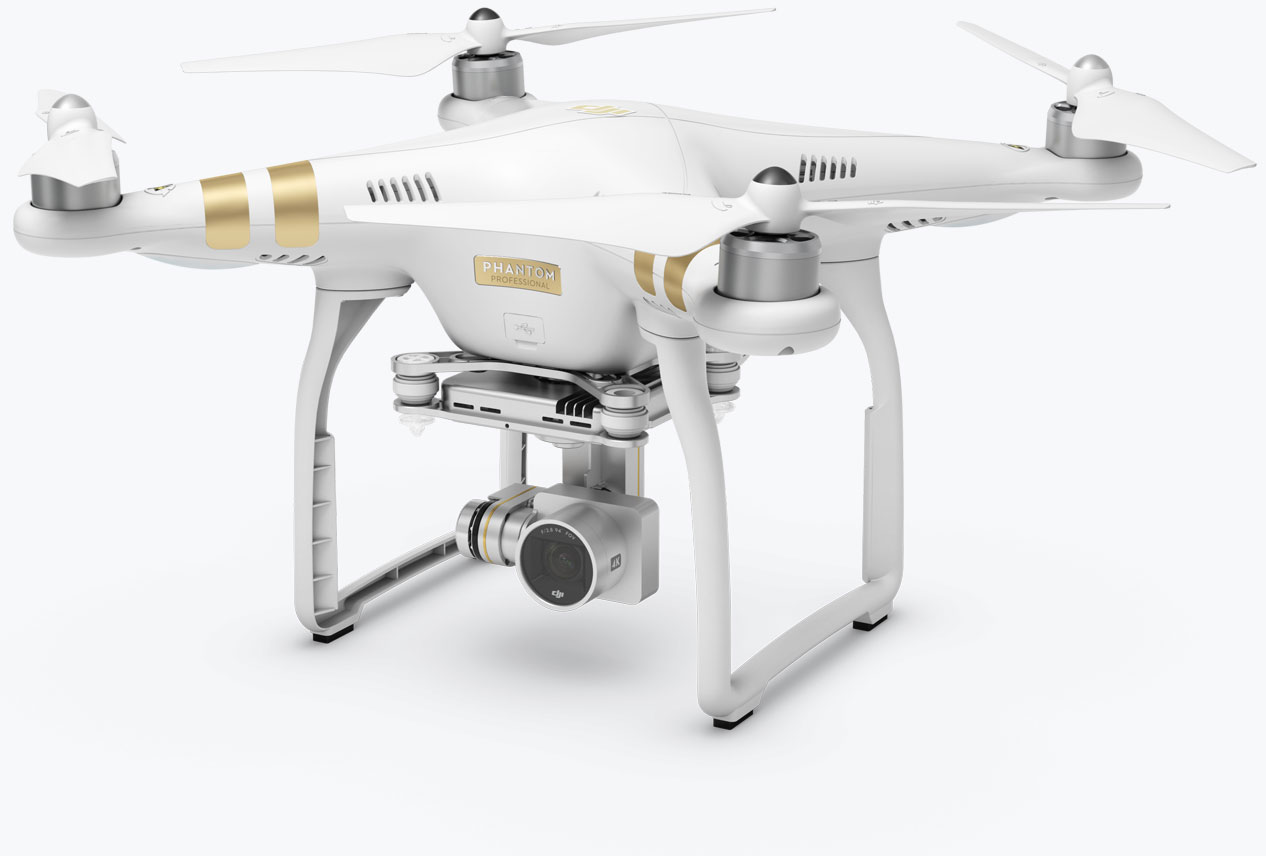}} & {\fontsize{10}{17}\selectfont \makecell{Phantom 3 Professional, DJI, 1280 gram, diagonal size=350mm, \bigstrut\\ Max speed=16m/s($\sim$57kph), max flight time is 23 minutes, \bigstrut\\ Flight time is reduced to 18 mins due to additional weight \cite{Phantom345:phantom3}.}}
\bigstrut \\
\hline
\Topspace
\Bottomspace
\multirow{3}{*}[0.4in]{\includegraphics[width=0.2\textwidth]{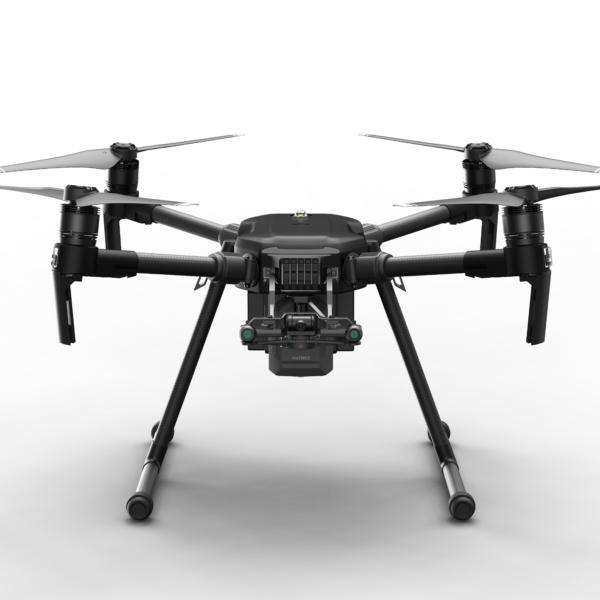}} & {\fontsize{10}{17}\selectfont \makecell{Matrice 200, DJI, 3.80kg, size:716mm $\times$ 220mm $\times$ 236mm,  \bigstrut\\ payload  up to 2kg, 16m/s ($\sim$61kph), batteries: (TB50) and TB55 \bigstrut\\ Max flight time: 38 minutes, operation range of 7km \cite{DJITheWo79:Matrice200}.}}
\bigstrut \\
\hline
\Topspace
\Bottomspace
\multirow{3}{*}[0.3in]{\includegraphics[width=0.2\textwidth, height=25mm]{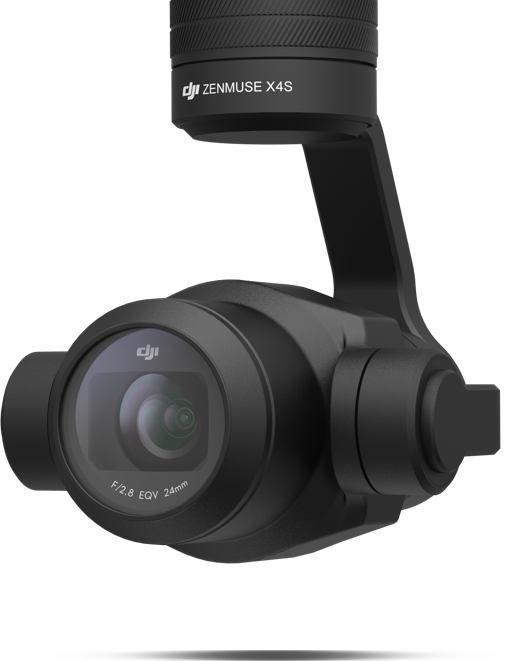}} & {\fontsize{10}{17}\selectfont \makecell{Zenmuse X4S, DJI, gimbal: Matrice 200, weight: 253 gram, \bigstrut \\ Fielf Of View (FOV): \ang{84}, resolution: Full HD to Cinematic 4K  \bigstrut \\ sensor: CMOS 20MPixels \cite{DJIZenmu25:zenmuse}.}}
\bigstrut \\
\hline
\Topspace
\Bottomspace
\multirow{3}{*}[0.25in]{\includegraphics[width=0.20\textwidth]{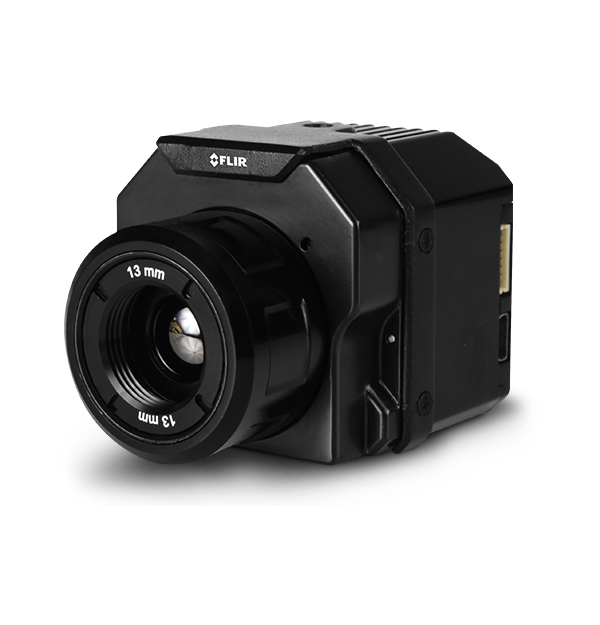}} & {\fontsize{10}{17}\selectfont \makecell{Vue Pro R, FLIR, IR camera, control: Bluetooth and  \bigstrut \\ Pulse Width Modulation (PWM) signal, FOV: \ang{45}, resolution: 640 $\times$ 512 \bigstrut \\ Lens: 6.8mm thermal, no gimbal \cite{FLIRVueP5:online}.}}
\bigstrut \\
\hline
\Topspace
\Bottomspace
\multirow{2}{*}[0.3in]{\includegraphics[width=0.20\textwidth]{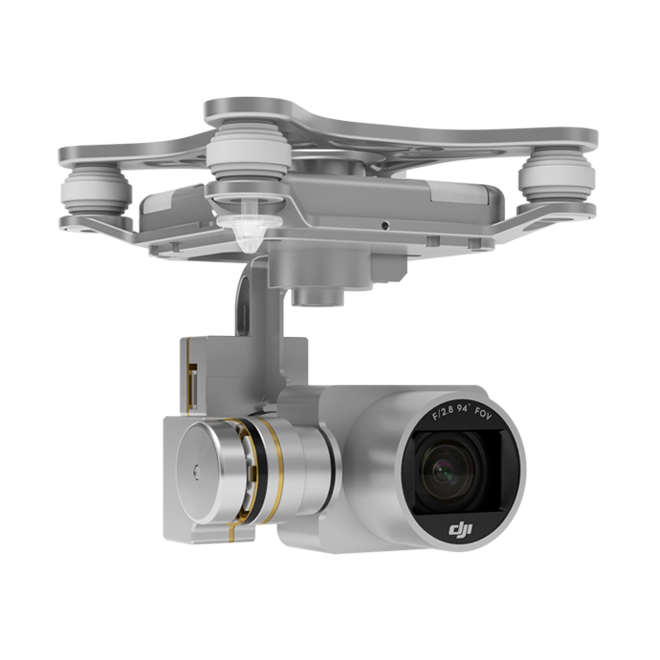}} & {\fontsize{10}{17}\selectfont \makecell{Phantom 3 camera, DJI, sensor: 1/2.3" CMOS 12.4MPixels \bigstrut \\  FOV: \ang{94}, (FPSs): 24 to 60, resolution: HD, FHD, UHD \cite{Phantom345:phantom3}.}}
\bigstrut \\
\bottomrule 
\end{tabular}
} % end of \resizebox{1\linewidth}{!}{
} % end of \centering{
% \vspace{-10pt}
\end{table*}
\begin{figure*}[bt]
	\centering
	\includegraphics[width=1\linewidth]{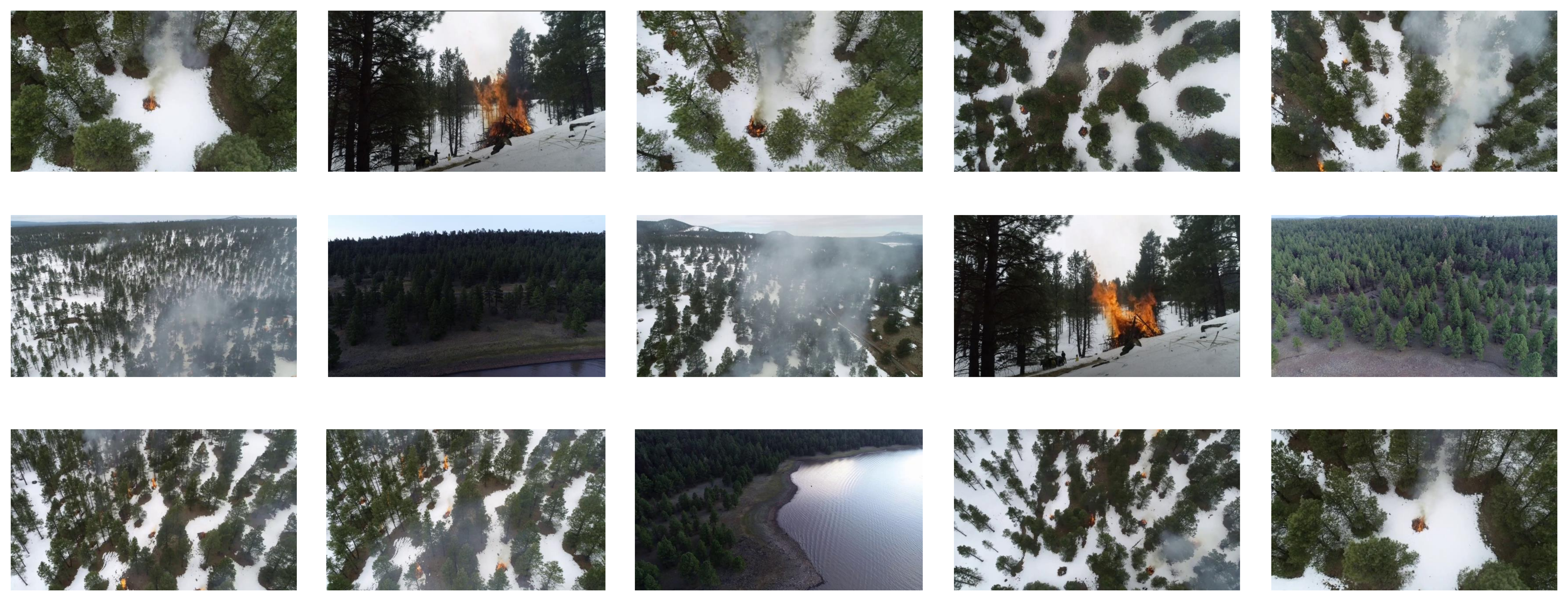}
	\caption{Frame samples of the normal spectrum palette.} 
% 	\vspace{-10pt}
    \label{fig:dataset_normal}
\end{figure*}
% *****************************************Figures

\subsection{Applicable data for the defined problem}
\label{subsec:Data}
This section presents the details of the %different types of
captured images, videos. 
% \as{I meant the nature of the data such as resolution, type of data, and how they were captured.}
%All information are captured in a video format for the sake of the simplicity. Then, the 
The captured videos are converted to frames based on the recorded Frames Per Second (FPS).
% \arr{What do you mean here: since videos are always formed of video frames, do you mean that we extract frames from the capture video for still-image processing?}. \as{Yes}
%Four type of videos including the normal spectrum palette, Fusion, WhiteHot, and GreenHot are available as a dataset.
Four types of video including the normal spectrum, fusion, white-hot, and green-hot palettes are available in the FLAME dataset \cite{qad6-r683-20}.

% \footnote{\url{https://github.com/AlirezaShamsoshoara/THELINKGOESHERE}}.

%*****************************************Figures
\begin{figure*}[bt]
	\centering
	\includegraphics[width=1\linewidth]{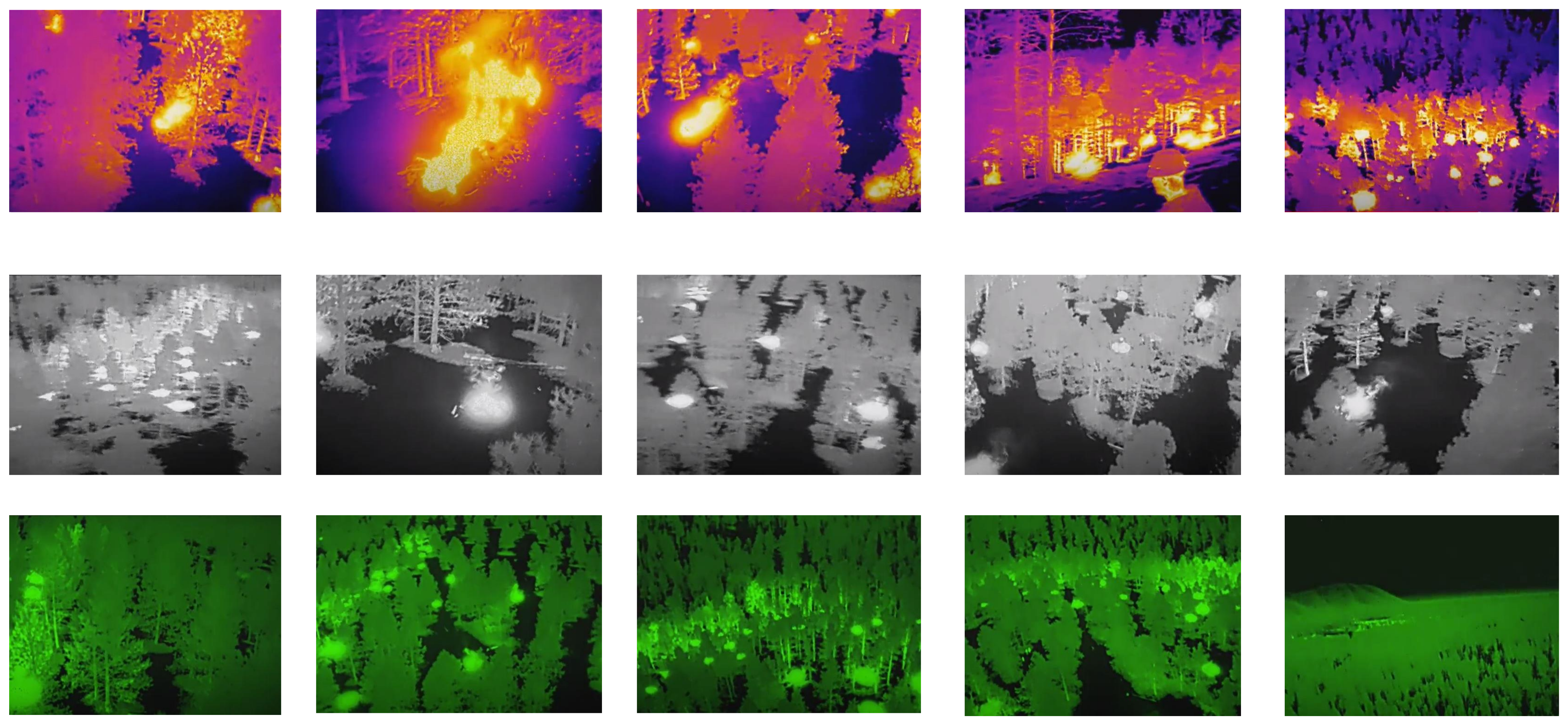}
	\caption{Frame samples of thermal images including Fusion, WhiteHot, and GreenHot palettes from top row to the bottom row.} 
% 	\vspace{-10pt}
    \label{fig:dataset_thermal}
\end{figure*}
% *****************************************Figures

The normal spectrum palette was recorded using both Zenmuse X4S and the phantom 3 camera. Other thermal and IR outputs were collected using the Forward Looking Infrared (FLIR) vue Pro R camera. 
%16 minutes video of both fire and no fire footage is available at the dimension of 1280 $\times$ 720 with 29 FPS. 
Several video clips which include both fire and no fire footage are available. The FLIR camera has a 1280 $\times$ 720 resolution with 29 fame per seconds (FPS).
Another 6 minutes of video is available for one pile burn from the start of the burning at 1280 $\times$ 720 resolution and 29 FPS. 
% Also, 6 minutes of No-Fire video is available at 1920 $\times$ 1080 resolution and 30 FPS.
The H.264 codec was used for all the recordings. More details about these videos are available in Table~\ref{tab:dataset_info} along with the dataset link. Figure~\ref{fig:dataset_normal} demonstrates some representative frames from both fire and no-fire videos. The full videos are available in the FLAME dataset repository.

% *****************************************Table
% \renewcommand{\arraystretch}{2.0}
\begin{table*}[bt]
\caption{Dataset information (\href{https://ieee-dataport.org/open-access/flame-dataset-aerial-imagery-pile-burn-detection-using-drones-uavs}{\blue{Link}}) \cite{qad6-r683-20}.}
 \centering{
\label{tab:dataset_info}
\resizebox{1.1\linewidth}{!}{  %fit to windows command
% \color{blue}
\renewcommand{\arraystretch}{2.5}
\begin{tabular}{c|c|c|c|c|c|c|c|c|c|c}
\toprule
\toprule
\Topspace
\Bottomspace
\textbf{\Large} & {\fontsize{12}{17}\selectfont \textbf{Type}} & {\fontsize{12}{17}\selectfont \textbf{Camera}} & {\fontsize{12}{17}\selectfont \textbf{Palette}} & {\fontsize{12}{17}\selectfont \textbf{Duration}} & {\fontsize{12}{17}\selectfont \textbf{Resolution}} & {\fontsize{12}{17}\selectfont \textbf{FPS}} & {\fontsize{12}{17}\selectfont \textbf{Size}} & {\fontsize{12}{17}\selectfont \textbf{Application}} & {\fontsize{12}{17}\selectfont \textbf{Usage}} & {\fontsize{12}{17}\selectfont \textbf{Labeled}}
\\
\hline
\Topspace
\Bottomspace
{
\fontsize{10}{17}\selectfont \textbf{1}} & \fontsize{12}{17}\selectfont Video & \fontsize{12}{17}\selectfont Zenmuse & 
\fontsize{12}{17}\selectfont Normal(.MP4) &
\fontsize{12}{17}\selectfont 966 seconds &
\fontsize{12}{17}\selectfont 1280$\times$720 &
\fontsize{12}{17}\selectfont 29 &
\fontsize{12}{17}\selectfont 1.2 GB &
\fontsize{12}{17}\selectfont Classification &
\fontsize{12}{17}\selectfont - &
\fontsize{12}{17}\selectfont N 
\\
\hline
\Topspace
\Bottomspace
{
\fontsize{10}{17}\selectfont \textbf{2}} & \fontsize{12}{17}\selectfont Video & \fontsize{12}{17}\selectfont Zenmuse & 
\fontsize{12}{17}\selectfont Normal(.MP4) &
\fontsize{12}{17}\selectfont 399 seconds &
\fontsize{12}{17}\selectfont 1280$\times$720 &
\fontsize{12}{17}\selectfont 29 &
\fontsize{12}{17}\selectfont 503 MB &
\fontsize{12}{17}\selectfont - &
\fontsize{12}{17}\selectfont - &
\fontsize{12}{17}\selectfont N 
\\
\hline
\Topspace
\Bottomspace
{
\fontsize{10}{17}\selectfont \textbf{3}} & \fontsize{12}{17}\selectfont Video & \fontsize{12}{17}\selectfont FLIR & 
\fontsize{12}{17}\selectfont WhiteHot(.MOV) &
\fontsize{12}{17}\selectfont 89 seconds &
\fontsize{12}{17}\selectfont 640$\times$512 &
\fontsize{12}{17}\selectfont 30 &
\fontsize{12}{17}\selectfont 45 MB &
\fontsize{12}{17}\selectfont - &
\fontsize{12}{17}\selectfont - &
\fontsize{12}{17}\selectfont N 
\\
\hline
\Topspace
\Bottomspace
{
\fontsize{10}{17}\selectfont \textbf{4}} & \fontsize{12}{17}\selectfont Video & \fontsize{12}{17}\selectfont FLIR & 
\fontsize{12}{17}\selectfont GreenHot(.MOV) &
\fontsize{12}{17}\selectfont 305 seconds &
\fontsize{12}{17}\selectfont 640$\times$512 &
\fontsize{12}{17}\selectfont 30 &
\fontsize{12}{17}\selectfont 153 MB &
\fontsize{12}{17}\selectfont - &
\fontsize{12}{17}\selectfont - &
\fontsize{12}{17}\selectfont N 
\\
\hline
\Topspace
\Bottomspace
{
\fontsize{10}{17}\selectfont \textbf{5}} & \fontsize{12}{17}\selectfont Video & \fontsize{12}{17}\selectfont FLIR & 
\fontsize{12}{17}\selectfont Fusion(.MOV) &
\fontsize{12}{17}\selectfont 25 mins &
\fontsize{12}{17}\selectfont 640$\times$512 &
\fontsize{12}{17}\selectfont 30 &
\fontsize{12}{17}\selectfont 2.83 GB &
\fontsize{12}{17}\selectfont - &
\fontsize{12}{17}\selectfont - &
\fontsize{12}{17}\selectfont N 
\\
\hline
\Topspace
\Bottomspace
{
\fontsize{10}{17}\selectfont \textbf{6}} & \fontsize{12}{17}\selectfont Video & \fontsize{12}{17}\selectfont Phantom & 
\fontsize{12}{17}\selectfont Normal(.MOV) &
\fontsize{12}{17}\selectfont 17 mins &
\fontsize{12}{17}\selectfont 3840$\times$2160 &
\fontsize{12}{17}\selectfont 30 &
\fontsize{12}{17}\selectfont 32 GB &
\fontsize{12}{17}\selectfont - &
\fontsize{12}{17}\selectfont - &
\fontsize{12}{17}\selectfont N 
\\
\hline
\Topspace
\Bottomspace
{
\fontsize{10}{17}\selectfont \textbf{7}} & \fontsize{12}{17}\selectfont Frame & \fontsize{12}{17}\selectfont Zenmuse & 
\fontsize{12}{17}\selectfont Normal(.JPEG) &
\fontsize{12}{17}\selectfont 39,375 frames &
\fontsize{12}{17}\selectfont 254$\times$254 &
\fontsize{12}{17}\selectfont - &
\fontsize{12}{17}\selectfont 1.3 GB &
\fontsize{12}{17}\selectfont Classification &
\fontsize{12}{17}\selectfont Train/Val &
\fontsize{12}{17}\selectfont Y 
\\
\hline
\Topspace
\Bottomspace
{
\fontsize{10}{17}\selectfont \textbf{8}} & \fontsize{12}{17}\selectfont Frame & \fontsize{12}{17}\selectfont Phantom & 
\fontsize{12}{17}\selectfont Normal(.JPEG) &
\fontsize{12}{17}\selectfont 8,617 frames &
\fontsize{12}{17}\selectfont 254$\times$254 &
\fontsize{12}{17}\selectfont - &
\fontsize{12}{17}\selectfont 301 MB &
\fontsize{12}{17}\selectfont Classification &
\fontsize{12}{17}\selectfont Test &
\fontsize{12}{17}\selectfont Y 
\\
\hline
\Topspace
\Bottomspace
{
\fontsize{10}{17}\selectfont \textbf{9}} & \fontsize{12}{17}\selectfont Frame & \fontsize{12}{17}\selectfont Phantom & 
\fontsize{12}{17}\selectfont Normal(.JPEG) &
\fontsize{12}{17}\selectfont 2,003 frames &
\fontsize{12}{17}\selectfont 3480$\times$2160 &
\fontsize{12}{17}\selectfont - &
\fontsize{12}{17}\selectfont 5.3 GB &
\fontsize{12}{17}\selectfont Segmentation &
\fontsize{12}{17}\selectfont Train/Val/Test &
\fontsize{12}{17}\selectfont Y(Fire) 
\\
\hline
\Topspace
\Bottomspace
{
\fontsize{10}{17}\selectfont \textbf{10}} & \fontsize{12}{17}\selectfont Mask & \fontsize{12}{17}\selectfont - & 
\fontsize{12}{17}\selectfont Binary(.PNG) &
\fontsize{12}{17}\selectfont 2,003 frames &
\fontsize{12}{17}\selectfont 3480$\times$2160 &
\fontsize{12}{17}\selectfont - &
\fontsize{12}{17}\selectfont 23.4 MB &
\fontsize{12}{17}\selectfont Segmentation &
\fontsize{12}{17}\selectfont Train/Val/Test &
\fontsize{12}{17}\selectfont Y(Fire) 
\\
\bottomrule 
\end{tabular}
} % end of \resizebox{1\linewidth}{!}{
} % end of \centering{
% \vspace{-10pt}
\end{table*}
% *****************************************Table
% \ar{The table text is barely visible, can you re-organize and use a larger font. and the links point to a wrong website?}
% \as{I did my best to enlarge it. Also, in the review mode there are less space from left and right, once the paper is accepted, it will be expanded more to left and right based on the final version limitation.}

The FLAME dataset also includes thermal videos such as Fusion, WhiteHot, and GreenHot palettes.
All videos were captured with the resolution of 640 $\times$ 512 and with 30 FPS. Multiple videos of fire and no-fire types with different lengths are available. 
Figure~\ref{fig:dataset_thermal} shows some randomly selected frames for these thermal videos. More details about the FLAME dataset are available in Table~\ref{tab:dataset_info}. Also, a sample video of this dataset is available on YouTube~\cite{youtube2020_dataset}. Sections~\ref{subsec:classification} and \ref{subsec:segmentation} demonstrate some of the videos conversions into frames to address research challenges such as fire classification and fire segmentation. Researchers can use applications of their choice to extract the frames from the videos based on the required FPS. The FLAME dataset including all images, videos, and data are available on IEEE-Dataport \cite{qad6-r683-20}.

\section{Goals: Suggested Experiments and Methodology}
\label{sec:method}
% This section studies three different approaches on the collected dataset to define the open challenges on this dataset. 
This section presents two example applications that can be defined based on the collected FLAME dataset along with Deep Learning solutions for these problems.
The first problem is the fire versus no-fire classification using a deep neural network (DNN) approach. The second problem deals with fire segmentation, which can be used %for the fire annotation and making a masked output from the camera's output. 
for fire detection by masking the identified fire regions on video frames classified as fire-containing in the first problem.

% The third problem concerns optimal UAV scheduling by developing the most efficient strategies for drones to frequently visit the pile fires distributed across the fire until they are fully burned while considering the available resources in terms of the number, type and available energy of the UAVs. \ar{This problem can be used to optimally cover pile and other spotted fires at minimum time under hardware constraints.}

\subsection{Fire vs No-Fire Classification}
\label{subsec:classification}
\begin{figure}[tp]
	\centering
	\includegraphics[height=0.95\textheight, keepaspectratio]{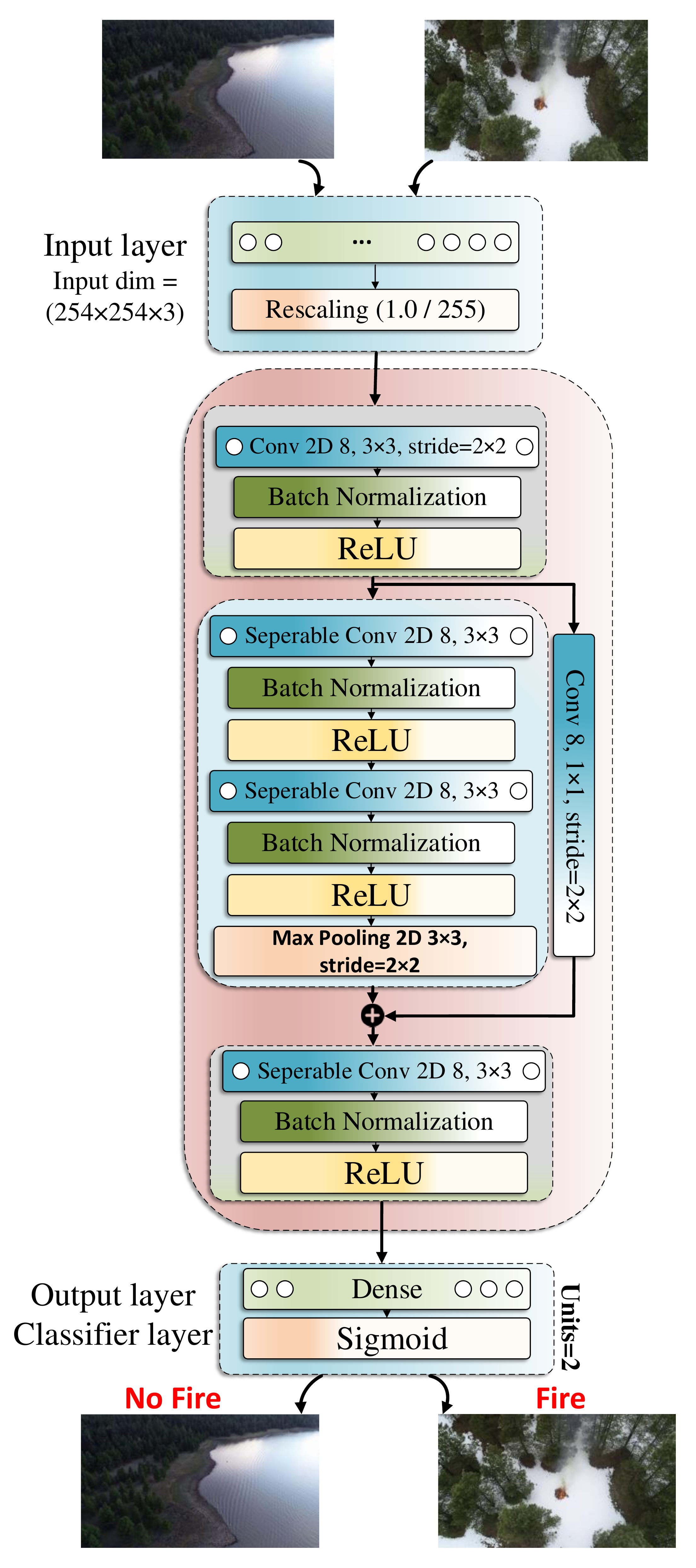}
	\caption{Small version of the Xception network for the fire classification.
% 	\ar{for non-fire, we can take two approaches, using forest images with no fire similar to the fire images, and using general photo dataset, and I think the former is more reasonable...} \as{We have already done the former one.}
} 
% 	\vspace{-10pt}
    \label{fig:Xception_model}
\end{figure}

The image classification problem is one of the challenging tasks in the image processing domain. In the past, traditional image processing techniques utilized RGB channel comparison to detect different objects such as fire in frames or videos \cite{celik2009fire, umar2017state, binti2015fire}. These traditional methods are not free of errors and are not fully reliable \cite{yuan2015uav}. For instance, RGB value comparison methods that usually consider a threshold value to detect fire may detect sunset and sunrise as a false positive outcome. However, training a DNN to perform this image classification task helps to learn elements not germane to the fire. Also, some studies such as \cite{qi2009computer, kundu2018highly} perform pixel-based classification and segmentation based on the HSV (Hue, Saturation, Value) format.
In the present study, a supervised machine learning method is used to classify the captured frames from camera.
% \ar{can we mention our policy for mixed images, when fire and non-fire parts coexist. how we label them?} \as{updated here:}
For mixed images when fire and non-fire parts coexist, the frame will be considered as the fire-labeled frame and when there is no fire in the frame, it will considered as non-fire-labaled. Instead of the green or fusion heat map, the normal range spectrum of images for the classification was selected using the Zenmuse X4S and the camera from DJI Phantom 3. The binary classification model which was used in this study is the Xception network \cite{chollet2017xception} proposed by Google-Keras\footnote{https://keras.io/examples/vision/image\_classification\_from\_scratch/}. The Xception model is a deep Convolutional Neural Network (DCNN). The structure of the DCNN is shown in Fig.~\ref{fig:Xception_model}. Replacing the standard \textit{Inception} modules of the Inception architecture with depth-wise separable convolutions resulted in the Xception network \cite{chollet2017xception, szegedy2016rethinking, szegedy2016inception}. 

%*****************************************Figures
% *****************************************Figures

Figure~\ref{fig:Xception_model} is the concise version of the Xception model. The Xception model has three main blocks: 1) the input layer, 2) the hidden layers, and 3) the output layer. The size of the input layer depends on the image size and the number of channels which in our case is ($254\times254\times3$). Then the value of RGBs in different channels are scaled to a float number between 0 and 1.0. The hidden layers rely on depth-wise separable convolutions and shortcut between the convolution blocks (ResNet~\cite{he2016deep}). The entry flow of the hidden layers is a pair of 2-Dimensional (2D) convolutional blocks with a size of 8 and a stride of $2\times2$. Each block follows a batch normalization and a Rectified Linear Unit (ReLU) activation function \cite{li2017convergence}. The batch normalization is used to speed up the training process and bring more randomness by decreasing the importance of initial weights and regularize the model. Next, the model follows two separable 2D convolutional blocks.
% and this procedure repeats for four times with another three times. However, the CNN size is different in each block \{128, 256, 512, 728\}
The last block of the hidden layer is a separable 2D convolutional layer with a size of 8 followed by another batch normalization and the ReLU function. Since the fire-detection is a binary classification task (Fire/No Fire), the activation function for the output layer is a Sigmoid function. The equation for the Sigmoid function is shown in (\ref{eq:sigmoid}), 
%\fbox{Razi Reviewed until here}.
% *****************************************Equation
\begin{align}\label{eq:sigmoid}
P(\textrm{label=Fire}) = \sigma(\textrm{label=Fire} | \zeta(\theta)) = \frac{1}{1 + e^{-\zeta(\theta)}}, 
\end{align}
% *****************************************Equation
where $\zeta(\theta)$ is the value of the output layer which is extracted based on the input frames and the RGB values of each pixel and all the weights across the hidden network. $\theta$ is the weight for the last layer of the network. The output value of the Sigmoid function is the probability of fire detection based on the imported frames into the network. To train the Xception 
% \arr{is this a typo?}
network and find the weights of all neurons, a value loss function is targeted to increase the accuracy of the networks and find the optimal values for the weights. As the problem in this section is a binary classification, the considered loss function is a binary cross-entropy defined as% . Equation (\ref{eq:lossfunc_bce}) expresses the binary cross-entropy, 
% *****************************************Equation
\begin{align}\label{eq:lossfunc_bce}
&\mathcal{L}(y, \hat{y}) = 
\\
\nonumber
&- \frac{1}{N} \sum\limits_{i=1}^N (y_{i} * \log(p(\hat{y}_{i})) + (1-y_{i}) * \log(1 - p(\hat{y}_{i})), 
\end{align}
% *****************************************Equation
where $N$ is the number of total samples in each batch used to update the loss function for each epoch. $y$ is the ground truth %truth ground 
label for the frames of types fire ($y=1$) and no/fire ($y=0$) %fire and no/fire frames 
based on the training data. 
%1 allocates to fire and 0 allocates to no-fire frames. 
$p(\hat{y})$ is the predicted probability of a frame belonging to the fire class. %the fire frame. 
Next, the Adam optimizer is used to minimize the loss function and find the optimal weights during the learning process. After training the network with the training dataset, the evaluation is performed using a test dataset in Section~\ref{sec:results}. The implemented code for this learning model is available on GitHub~\cite{github:code_dataset_fire}.
% \arr{Better to make a reference instead of footnote}\footnote{\url{https://git.io/Jk838}}.

\subsection{Fire Segmentation}
\label{subsec:segmentation}
\begin{figure}[tp]
	\centering
	\includegraphics[height=0.95\textheight, keepaspectratio]{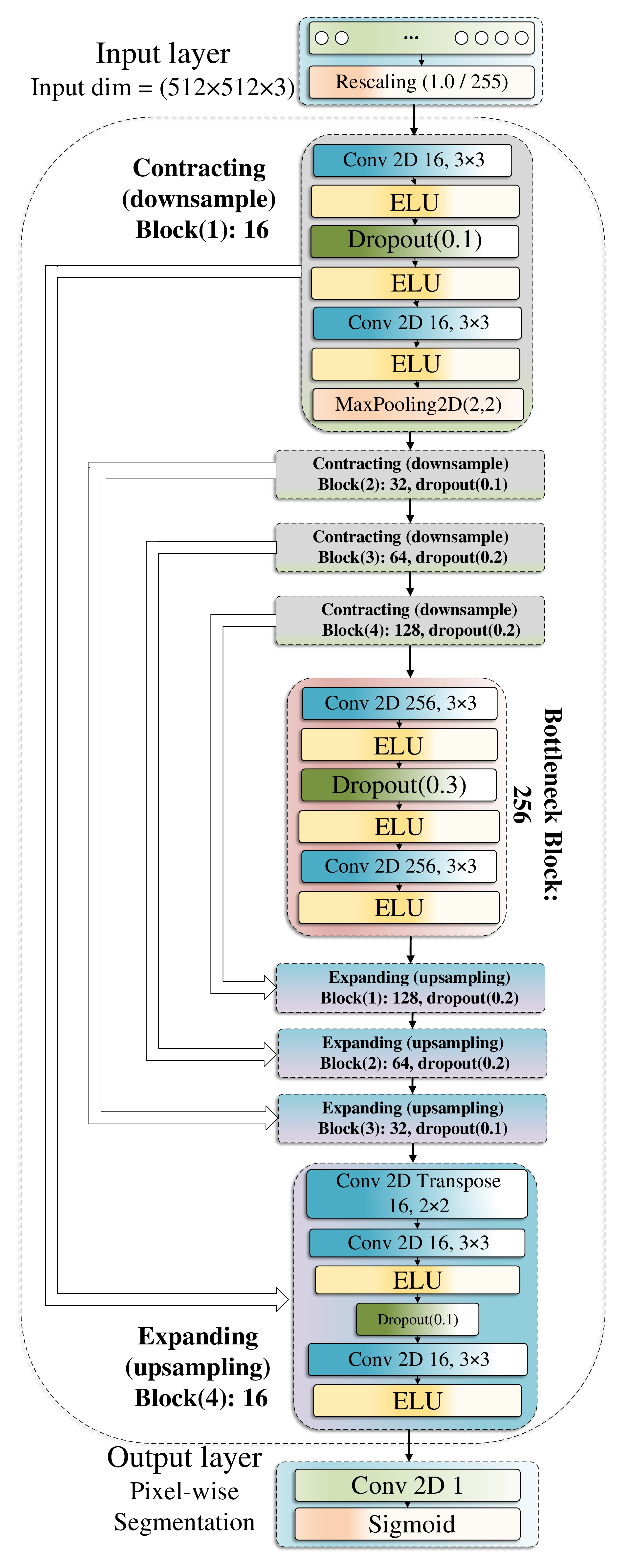}
	\caption{Customized version of the U-Net for the fire segmentation.} 
% 	\vspace{-10pt}
    \label{fig:customized_unet}
\end{figure}

This section considers the problem of image segmentation for frames labeled as "fire" by the fire classification algorithm presented in section \ref{subsec:classification}. Studying the fire segmentation problem is useful for scenarios like detecting small fires \cite{yuan2017aerial}. Also, fire segmentation helps fire managers localize different discrete places of active burning for the purpose of fire monitoring. The goal is to propose an algorithm to find the pile burn segments in each frame and generate relevant masks. These segmentation problems were handled differently in the past using image processing and RGB threshold values to segment different data batches 
% \arr{error-free methods wont exist so better to say}
which exhibits relatively high error rates \cite{ccelik2007fire, yuan2015uav, khalil2020fire}. The goal is to develop an image semantic segmentation to perform a pixel-wise classification for each frame at the pixel level to define a fire mask for the generated output. To accomplish this task, a DCNN model is implemented to predict the label of each pixel based on the imported data. This segmentation problem can be recast as a binary pixel-wise classification problem, where each pixel can take two labels: "fire" and "non-fire" (background). To accomplish the image segmentation task, the fire test dataset from Section~\ref{subsec:classification} is considered as a training dataset. To train a DCNN model, a Ground Truth Mask dataset is required. Different tools and applications such as Labelbox~\cite{Labelbox:online}, Django Labeller~\cite{Django:online}, LabelImg~\cite{labelImg:online}, MATLAB Image Labeler~\cite{MATLABImageLabeler:online}, GNU Image Manipulation Program (GIMP)~\cite{GIMPGNUI35:online}, etc are available to perform different types of the manual image segmentation such as pixel labeling, annotation (rectangles, lines, and cuboid) on the Regions Of Interest (ROI) to provide training data for the utilized deep learning model. The MATLAB (TM) Image Labeler is used on 2003 frames to generate the Ground Truth Masks. This subcategory of the FLAME dataset of masks and images is presented in Table~\ref{tab:dataset_info}. The implemented image segmentation model is adopted from the U-Net convolutional network developed for biomedical image segmentation \cite{ronneberger2015u}. U-Net is an end-to-end technique between the raw images and the segmented masks. A few changes are made to this network to accommodate the FLAME dataset and adapt it to the nature of this problem. The ReLU activation function is changed to Exponential Linear Unit (ELU) of each two-dimensional convolutional layer to obtain more accurate results \cite{ELU:online}. The ELU function has a negative outcome smaller than a constant $\alpha$ for the negative input values and it exhibits a smoother behavior than the ReLU function. The structure of the customized U-Net is shown in Figure~\ref{fig:customized_unet}. The backbone of the U-Net consists of a sequence of up-convolutions and concatenation with high-resolution features from the contracting path. 

The size of the input layer is 512 $\times$ 512 $\times$ 3 designed to match the size of the inputs images and three RGB channels. For computational convenience, the RGB values (between 0 and 255) are scaled down by 255 to yield float values between 0 and 1. Next, it follows the first contracting block including a two-dimensional fully convolutional layers with the ELU activation function, a dropout layer, another same fully convolutional layer,
% \arr{do you mean: another fully convolutional layer with the same structure?} Alireza: Yes, based on the structure in the Fig.
and a two-dimensional 
% \arr{2 dimensional: always better to use 2-dimensional or two-dimensional to read as a word. I fixed most of them but please make sure we didn't miss any of them}
max pooling layer. This structure is repeated another three times to shape the left side of the U shape. Next, there are two two-dimensional fully connected layers with a dropout layer in between, the same structure of the left side is repeated for the right side of the U shape to have a symmetric structure for the up-convolution path in each block. Also, there exists a concatenation between the current block and the peer block from the contracting path. Since the pixel-wise segmentation is a binary classification problem, the last layer has the Sigmoid activation function.
% Razi revised slightly: The input layer considers the size of the targeted image which is 512 $\times$ 512 and the number of RGB channels which is 3 in this study (512 $\times$ 512 $\times$ 3). Since all RGB values of the data are between 0 and 255, we divide all values by 255 to have a float value between 0 and 1. Next, it follows the first contracting block including a 2 dimensional fully convolutional layers with the ELU activation function, a dropout layer, another same fully convolutional layer, and a 2 dimensional max pooling layer. This structure is repeated another three times to shape left side of the U shape. Next, there are two 2 dimensional fully connected layers with a dropout layer in between, the same structure of the left side is repeated for the right side of the U shape to have a symmetric structure for the up-convolution path plus in each block, there is a concatenation between the current block and the peer block from the contracting path. Since the pixel-wise segmentation is a binary classification problem, the last layer has the Sigmoid activation function.

%*****************************************Figures
% *****************************************Figures

The DCNN utilizes a dropout method to avoid the overfitting issue in the FLAME dataset analysis and realize a more efficient regularization noting the small number of ground truth data samples. %Dropout is a known concept: This approach is feasible by randomly drop the neural nodes during the training process over the same dataset. 
The utilized loss function is the binary cross entropy similar to (\ref{eq:lossfunc_bce}). The Adam optimizer is used to find the optimal value of weights for the neurons. The evaluation of the FLAME-trained model with the ground truth data is described in Section~\ref{subsec:res_segmentation}. The implemented code for this section is available on GitHub~\cite{github:code_dataset_fire}.

\section{Results: Metrics and guidance on reporting results}
\label{sec:results}
In this section, we present the results of the two different problems of fire classification and fire segmentation. First, we provide the details of the parameters used in our experiments. Next, we discuss the results of each algorithm. All simulations for the training, validation, and testing phases, are performed using a AMD Ryzen 9 3900X with NVidia RTX 2080 Ti on an Ubuntu system.  

%Razi revised slightly: This section brings the results for the three different challenges of the fire classification, fire segmentation and UAV scheduling for the piles observation. First, each section explained the parameters used in the evaluation and the simulation. Next, the results are discussed based on the methodology proposed in Section~\ref{sec:method}. All simulations and evaluations including training, validation, testing, were performed using a AMD Ryzen 9 3900X with NVidia RTX 2080 Ti on an Ubuntu system.  

\subsection{Fire vs No-Fire Classification}
\label{subsec:res_classification}
In the training section, the total number of the frames is 39,375 which includes 25,018 frames of type "fire" and 14,357 frames of type "non-fire".
%\arr{labeled creating an imbalanced dataset for the training phase-> if we don't tackle the imbalanced number of two classes by data augmentation, etc no need to highlight it here.} 
The training dataset is further split to 80\% training and 20\% validation sets. All frames are shuffled before feeding into the network. Also, augmentation methods such as horizontal flipping and random rotation are used to create new frames and address the issue of bias for unbalanced number of samples in the two "fire" and "non-fire" classes. The training phase ran over 40 epochs and the learning rate for the Adam optimizer is set to 0.001 which remains fixed during the training phase. Also, the batch size of 32 is used to fit the model in the training phase. 
To evaluate the accuracy and loss on the test dataset, 8,617 frames including 5,137 fire-labeled frames and 3,480 No-fire-labeled frames are fed into the pre-trained networks. 
Table~\ref{tab:accuracy_loss} reports loss and accuracy on training, validation, and test sets. It is noteworthy to mention that all frames for the training phase are collected using the Matrice 200 drone using Zenmuse X4S camera and all frames for the test set are collected using the Phantom drone and its default mounted camera. Therefore, no overlap exists between the training and test samples. This fact confirms that our method is not biased to the imaging equipment properties, and the actual accuracy would be even higher when using the same imaging conditions for the training and test phase. The achieved accuracy % \arr{of what: image classification?}
of the ``Fire vs No-Fire" classification is 76.23\%. 

\begin{figure}[bt]
	\centering
	\includegraphics[width=1\linewidth]{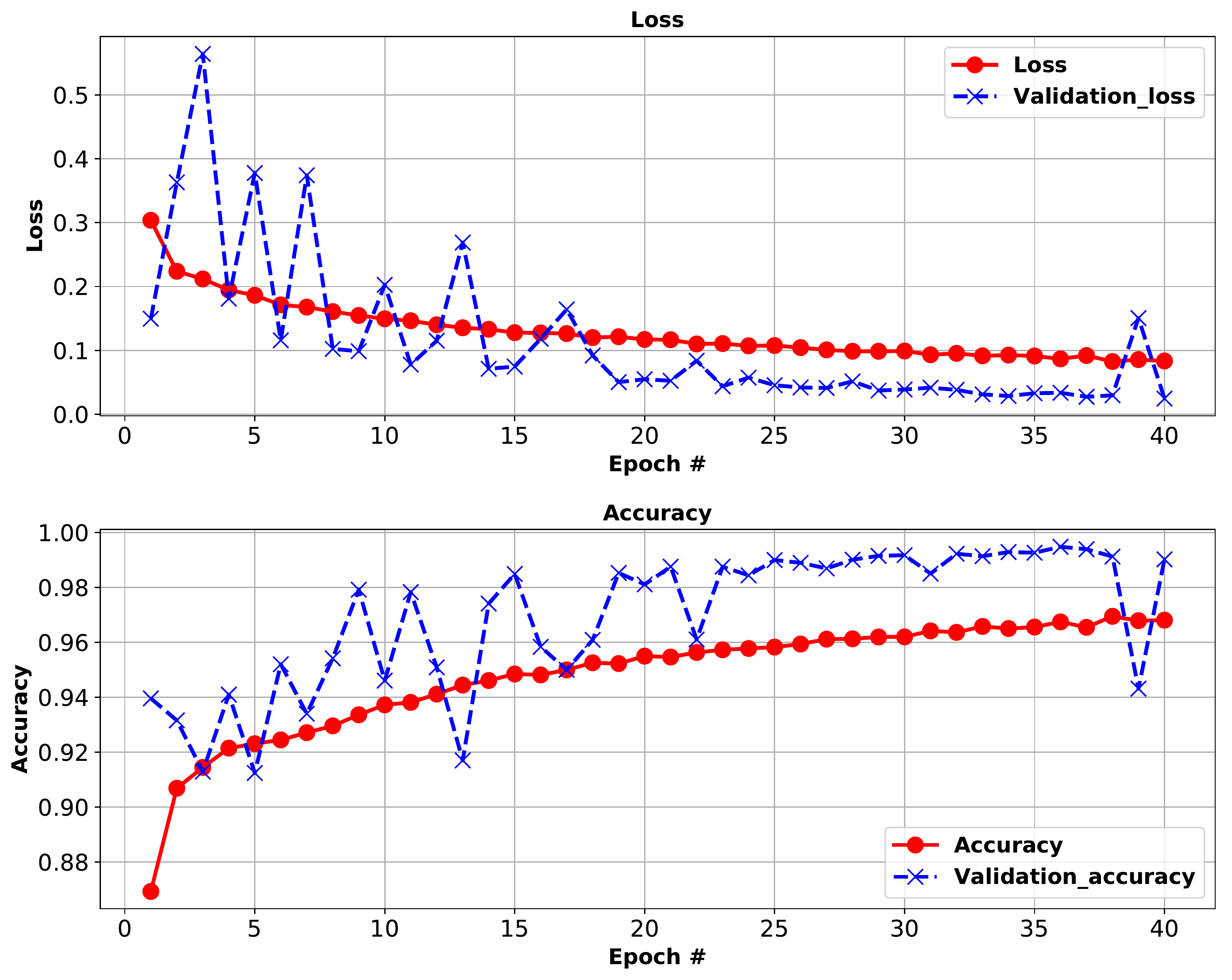}
	\caption{Accuracy and loss values for the training and validation sets.} 
% 	\vspace{-10pt}
    \label{fig:accuracy_loss_training}
\end{figure}

\begin{figure}[bt]
	\centering
	\includegraphics[width=1\linewidth]{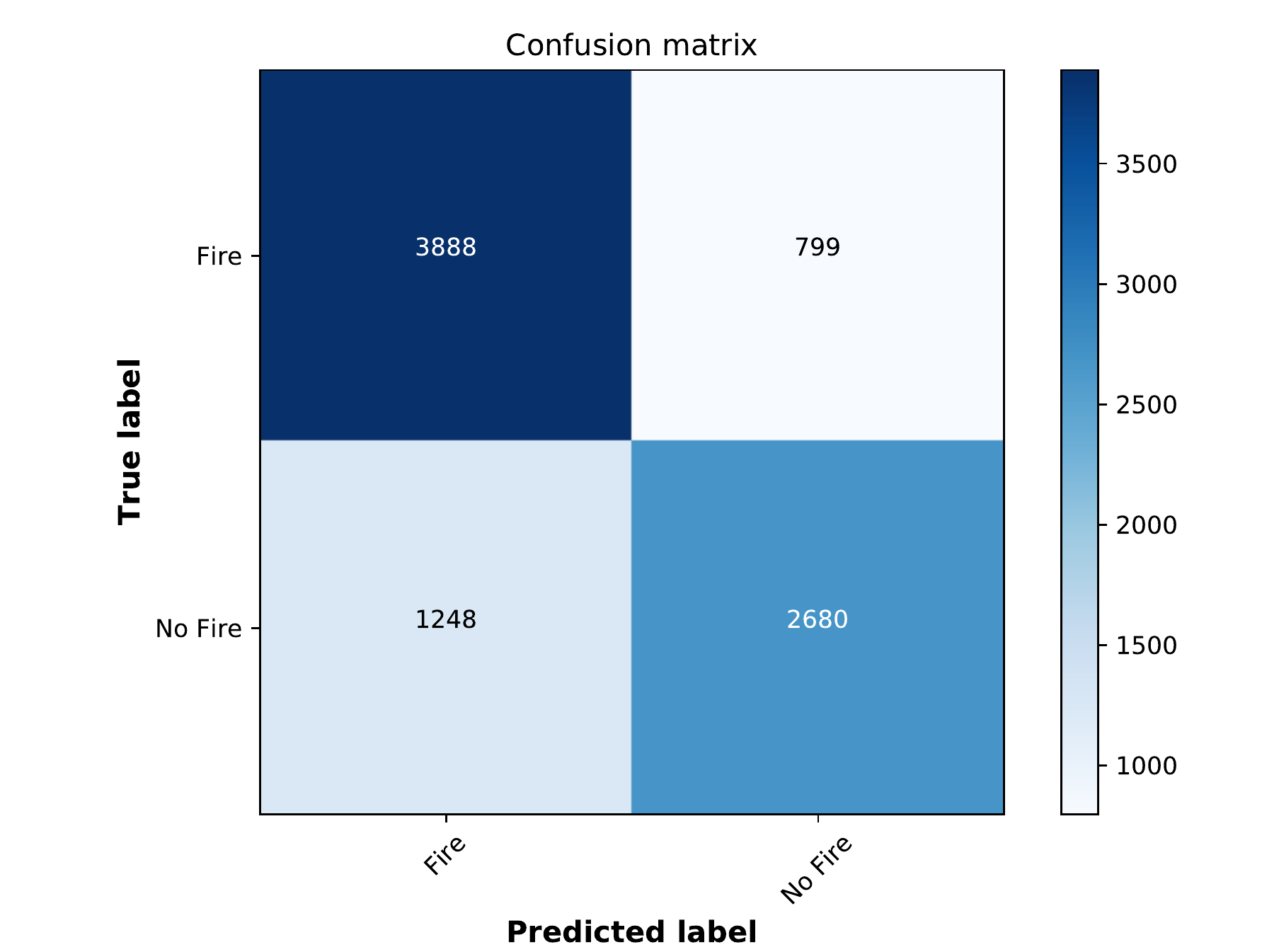}
	\caption{Confusion matrix for the true and predicted labels.} 
% 	\vspace{-10pt}
    \label{fig:Confusion_matrix}
\end{figure}

Figure~\ref{fig:accuracy_loss_training} demonstrates the loss and accuracy for the training phase for both the training and validation sets. Also, Figure~\ref{fig:Confusion_matrix} presents the confusion matrix for this binary fire classification task for all predictions. The vertical axis shows the true label of frames and the horizontal axis expresses the predicted label. The confusion matrix considers two classes which is plotted for the test dataset. Since, the ratio of fire and No-fire frames was imbalanced at the training phase, the rate of the false positive (classifying a true no-fire as fire) is higher than the false negative rate (classifying a true fire as no-fire).
% *****************************************Figures

%*****************************************Figures
% *****************************************Figures

% *****************************************Table
\begin{table}[bt]
\caption{Accuracy and loss for evaluation of the fire classification.}
 \centering{
\label{tab:accuracy_loss}
\resizebox{0.6\linewidth}{!}{  %fit to windows command
% \color{blue}
% \renewcommand{\arraystretch}{0.0}
\begin{tabular}{c|c|c}
\toprule
\toprule
\Topspace
\Bottomspace
\textbf{\Large} & \multicolumn{2}{c}{\textbf{\Large Performance}}
\\
\hline
\toprule
\toprule
\Topspace
\Bottomspace
\textbf{\large Dataset} & {\large Loss} & {\large Accuracy(\%)}
\\
\hline
\Topspace
\Bottomspace
{\fontsize{14}{17}\selectfont Test set} & \textbf{\fontsize{14}{17}\selectfont $0.7414$} & \textbf{\fontsize{14}{17}\selectfont $76.23$}
\\
\hline
\Topspace
\Bottomspace
{\fontsize{14}{17}\selectfont Validation set} & \fontsize{14}{17}\selectfont $0.1506$ & \fontsize{14}{17}\selectfont $94.31$
\\
\hline
\Topspace
\Bottomspace
{\fontsize{14}{17}\selectfont Training set} & \fontsize{14}{17}\selectfont $0.0857$ & \fontsize{14}{17}\selectfont $96.79$
\\
\bottomrule 
\end{tabular}
} % end of \resizebox{1\linewidth}{!}{
} % end of \centering{
% \vspace{-10pt}
\end{table}
% *****************************************Table

\subsection{Fire Segmentation}
\label{subsec:res_segmentation}
\begin{figure*}[bt]
	\centering
	\includegraphics[width=1\linewidth]{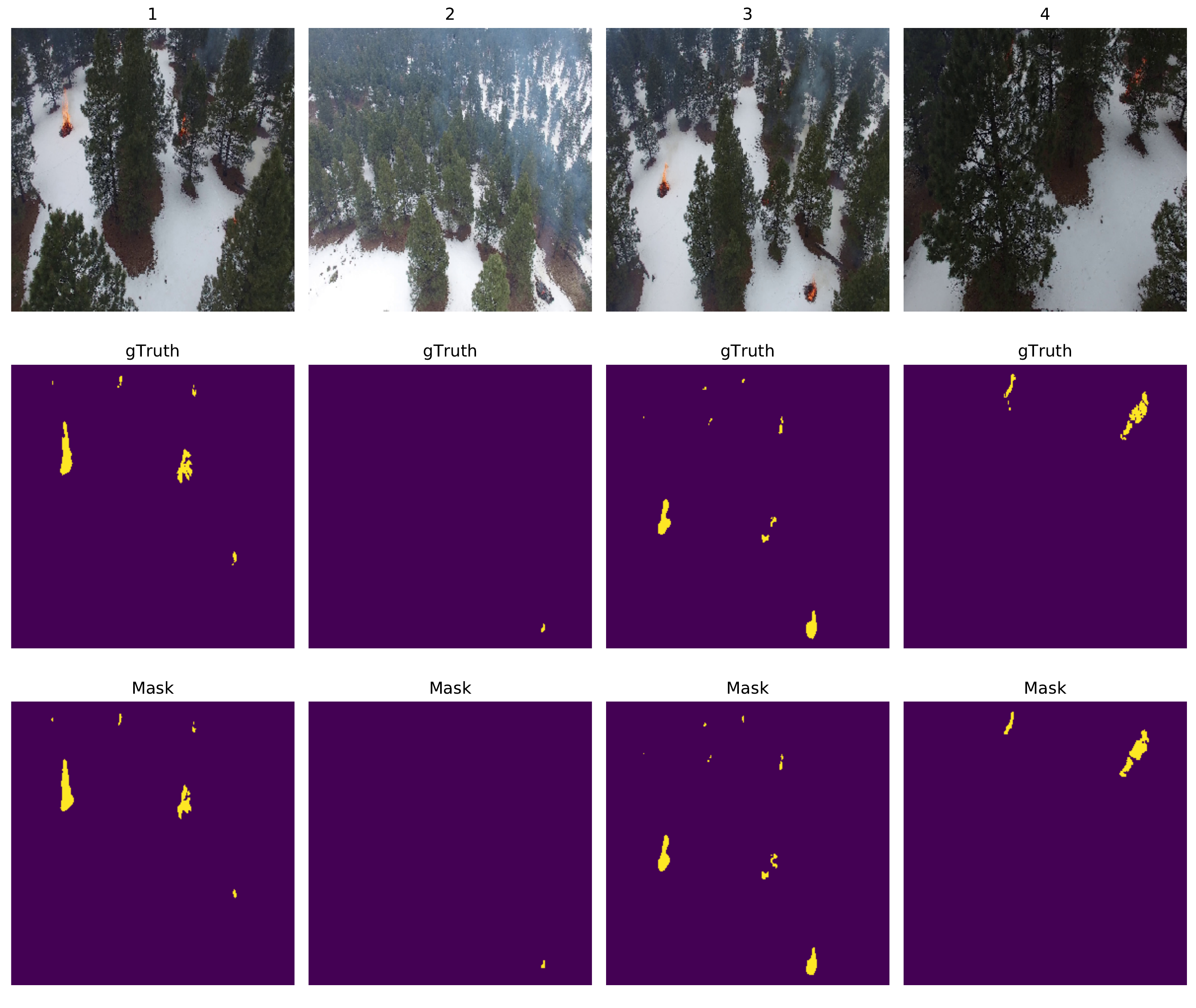}
	\caption{Performance of the fire segmentation on six frames of the test set.} 
% 	\vspace{-10pt}
    \label{fig:segmentation_test}
\end{figure*}

The purpose of fire segmentation is to accurately localize and extract the fire regions from the background. Therefore, the video frames within the test set which are labeled as "fire" by the fire classification stage in section (Section~\ref{subsec:classification}) are used here for training, validation, and test. The total number of frames is 5,137 and 2003 masks generated using the MATLAB (TM) Image Labeler tool for the ground truth data. The ground truth masks and data are generated based on the human subject matter expert (SME) eye efficiency to mark the fire pixels using manual polygon shape in MATLAB (TM) Image Labeler. The split ratio between the training and validation data is 85\% and 15\%. The frames and ground truth data were shuffled accordingly before importing into the training model. The maximum number of epochs is 30; however, an early stop callback was considered when the performance does not substantially change. The batch size for the training is 16. Figure~\ref{fig:segmentation_test} demonstrates six samples of the test set along with the expected ground truth masks and the generated masks from the trained network. The first row is the input frame to the model, the second row is the ground truth (gTruth) which is the expected mask, and the last row is the generated mask by the trained model. Also, Table~\ref{tab:segmentation_performance} shows the performance evaluation of this model. In this table, precision, recall, and Area Under Curve (AUC), F1-score, sensitivity, specificity, and Mean Intersection-Over-Union (Mean IOU) are reported.

%Razi revised slightly: Since the purpose of fire segmentation is to extract the fire, the fire labeled test set from the fire classification section (Section~\ref{subsec:classification}) is considered for the training, validation, and test data. The total number of frames is 5,137 and 2003 masks generated using the MATLAB Image Labeler tool for the ground truth data. The ground truth masks and data are generated based on the human eye efficiency to mark the fire pixels using manual polygon shape in MATLAB Image Labeler. The split ratio between the training and validation data is 85\% and 15\%. The frames and ground truth data were shuffled accordingly before importing to the training model. The maximum number of epochs is 30; however, an early stop callback was considered in case that the performance does not change. The batch size for the training is 16. Figure~\ref{fig:segmentation_test} demonstrates six samples of the test set along with the expected ground truth masks and the generated masks from the trained network. The first row is the input frame to the model, the second row is the ground truth (gTruth) which is the expected mask, and the last row is the generated mask by the trained model. Also, Table~\ref{tab:segmentation_performance} shows the performance evaluation of this model. In this table, precision, recall, and Area Under Curve (AUC), F1-score, sensitivity, specificity, and Mean Intersection-Over-Union (Mean IOU) are reported.

%*****************************************Figures
% *****************************************Figures

% *****************************************Table
\begin{table}[bt]
\caption{Performance evaluation of the customized U-Net on the fire dataset for the fire segmentation.}
 \centering{
\label{tab:segmentation_performance}
\resizebox{1\linewidth}{!}{  %fit to windows command
% \color{blue}
\begin{tabular}{c|c|c|c|c|c|c|c}
\toprule
\toprule
\Topspace
\Bottomspace
\textbf{\Large} & \multicolumn{7}{c}{\textbf{\Large Performance evaluation}}
\\
\hline
\toprule
\toprule
\Topspace
\Bottomspace
\textbf{\large Dataset} & {\large Precision(\%)} & {\large Recall(\%)} & {\large AUC(\%)} & {\large F1-Score(\%)} & {\large Sensitivity(\%)} & {\large Specificity(\%)} & {\large IOU(\%)}
\\
\hline
\Topspace
\Bottomspace
{\fontsize{14}{17}\selectfont Image Segmentation} & \textbf{\fontsize{14}{17}\selectfont $91.99$} & \textbf{\fontsize{14}{17}\selectfont $83.88$} & \textbf{\fontsize{14}{17}\selectfont $99.85$} &
\textbf{\fontsize{14}{17}\selectfont $87.75$} &
\textbf{\fontsize{14}{17}\selectfont $83.12$} &
\textbf{\fontsize{14}{17}\selectfont $99.96$} &
\textbf{\fontsize{14}{17}\selectfont $78.17$}
\\
\bottomrule 
\end{tabular}
} % end of \resizebox{1\linewidth}{!}{
} % end of \centering{
% \vspace{-10pt}
\end{table}

\section{Open Challenges regarding the dataset}
\label{sec:open_challenges}
This study proposed two different challenges regarding the dataset.
We encourage other researchers to consider this available FLAME dataset and improve the accuracy of the fire classification problem which might include providing more ground truth data as a labeled mask for the fire segmentation. 
Also, three other thermal images such as GreenHot, WhiteHot, and the fusion are also available for further investigation regarding the fire segmentation and classification.
Other considerations include the type of fire elements including the different structures of the fire (white hot core, exterior, etc). These elements can be segmented as different parts of the fire to have better understanding of the fire behavior.
Another challenge or problem could be investigating different fire detection models on these thermal images to see which type of the data has better accuracy for the model. Perhaps, another important research direction would be developing integrative imagery-based fire spread models by incorporating other environmental factors such as the  terrain model, and the vegetation fuel profile of the region. Extracting such factors from the images and videos and comparing with alternative sources can advance the image-based fire modeling algorithms. Other open challenges and future directions regarding the FLAME dataset include but not limited to 1) transfer learning, 2) context-based fire detection using a model and then zero-shot learning, 3) fire content analysis, 4) temporal analysis, 5) surrogate airborne perspective analysis, 6) metric design, 7) performance standards, 8) user displays, 9) edge node efficiency, and 10) occlusion robustness. 

%Razi revised slightly: This study proposed two different challenges regarding the dataset. We encourage other researchers to consider this available dataset and improve the accuracy of the fire classification problem and also providing more ground truth data as a labeled mask for the fire segmentation. Also, three other thermal images such as GreenHot, WhiteHot, and the fusion are also available for further investigation regarding the fire segmentation and classification. Another challenge or problem could be investigating different fire detection models on these thermal images to see which type of the data has better accuracy for the model. 

\section{Conclusion}
\label{sec:conclusion}
This paper provided the FLAME (Fire Luminosity Airborne-based Machine learning Evaluation) dataset for pile burns in Northern Arizona forest. Two drones were used to collect aerial frames and videos in four different palettes of normal, Fusion, WhiteHot, and GreenHot using normal and thermal cameras. The frames were used in two different applications, in the first challenge, a convolutional neural network was used as a deep learning binary fire classification to label data. In the second approach, a machine learning approach was proposed to extract fire masks from fire labeled data as an image segmentation technique. These exemplary applications show the utility of the FLAME dataset in developing computer tools for fire management and control. Also, FLAME dataset can be used as a benchmark dataset for testing generic image processing algorithms.
We provide numerical result for the performance of the proposed two algorithms developed for image classification and detection. We believe that developing more advanced models by the research community can further improve the reported results. Another potential use for this dataset is developing fire classification and detection algorithms by a collective analysis of different imaging modalities including regular and thermal images. Also, researchers can utilize fire segmentation methods to define related networking and monitoring problems, such as optimal task scheduling for a fleet of drones to optimally cover the pile burns in a certain region at shortest time possible.

%Razi revised slightly: This paper provided a dataset for pile burns in Northern Arizona forest. Two drones were used to collect aerial frames and videos in four different palettes of normal, Fusion, WhiteHot, and GreenHot using normal and thermal cameras. The frames were used in two different applications, in the first challenge, a convolutional neural network was used as a deep learning binary fire classification to label data. In the second approach, a machine learning approach was proposed to extract fire masks from fire labeled data as an image segmentation technique. The result section evaluates the performance of these three challenges and techniques. The future study of this dataset includes proposing fire segmentation and fire-NoFire classification on thermal images as well. Also, researchers can utilize fire segmentation methods to define UAV scheduling application such as the salesman problem to monitor the behaviour of burning piles.
\section*{Acknowledgment}
This material is based upon work supported by the Air Force Office of Scientific Research under award number FA9550-20-1-0090 and the National Science Foundation under Grant Number 2034218. Thanks to Neil Chapman and Paul Summerfelt of the Flagstaff Fire Department for providing access to the prescribed fire.

\bibliography{mybibfile}

\begin{thebibliography}{10}
\expandafter\ifx\csname url\endcsname\relax
  \def\url#1{\texttt{#1}}\fi
\expandafter\ifx\csname urlprefix\endcsname\relax\def\urlprefix{URL }\fi
\expandafter\ifx\csname href\endcsname\relax
  \def\href#1#2{#2} \def\path#1{#1}\fi

\bibitem{National39:online}
{National Interagency Fire Center},
  \url{https://www.nifc.gov/fireInfo/nfn.htm}, (Accessed on 10/28/2020).

\bibitem{National18:online}
{National Interagency Fire Center},
  \url{https://www.nifc.gov/fireInfo/fireInfo\\ \_statistics.html}, (Accessed
  on 10/28/2020).

\bibitem{toledo2018forest}
J.~Toledo-Castro, P.~Caballero-Gil, N.~Rodr{\'\i}guez-P{\'e}rez,
  I.~Santos-Gonz{\'a}lez, C.~Hern{\'a}ndez-Goya, R.~Aguasca-Colomo, {Forest
  fire prevention, detection, and fighting based on fuzzy logic and wireless
  sensor networks}, Complexity 2018 (2018) 1--17.
\newblock \href {http://dx.doi.org/10.1155/2018/1639715}
  {\path{doi:10.1155/2018/1639715}}.

\bibitem{kaur2019fog}
H.~Kaur, S.~K. Sood, {Fog-assisted IoT-enabled scalable network infrastructure
  for wildfire surveillance}, Journal of Network and Computer Applications 144
  (2019) 171--183.

\bibitem{coen2018transforming}
J.~L. Coen, W.~Schroeder, S.~D. Rudlosky, Transforming wildfire detection and
  prediction using new and underused sensor and data sources integrated with
  modeling, in: Handbook of Dynamic Data Driven Applications Systems, Springer,
  2018, pp. 215--231.

\bibitem{afghah2020cooperative}
F.~Afghah, A.~Shamsoshoara, L.~L. Njilla, C.~A. Kamhoua, Cooperative spectrum
  sharing and trust management in iot networks, Modeling and Design of Secure
  Internet of Things (2020) 79--109.

\bibitem{kamhoua2020modeling}
C.~A. Kamhoua, L.~Njilla, A.~Kott, S.~Shetty, Modeling and design of secure
  internet of things (2020).

\bibitem{huang2020wildfire}
Q.~Huang, A.~Razi, F.~Afghah, P.~Fule, {Wildfire Spread Modeling with Aerial
  Image Processing}, in: 2020 IEEE 21st International Symposium on" A World of
  Wireless, Mobile and Multimedia Networks"(WoWMoM), IEEE, 2020, pp. 335--340.

\bibitem{friedlingstein2019global}
P.~Friedlingstein, M.~Jones, M.~O'sullivan, R.~Andrew, J.~Hauck, G.~Peters,
  W.~Peters, J.~Pongratz, S.~Sitch, C.~Le~Qu{\'e}r{\'e}, et~al., {Global carbon
  budget 2019}, Earth System Science Data 11~(4) (2019) 1783--1838.

\bibitem{keshavarz2020real}
M.~Keshavarz, A.~Shamsoshoara, F.~Afghah, J.~Ashdown, A real-time framework for
  trust monitoring in a network of unmanned aerial vehicles, in: IEEE INFOCOM
  2020-IEEE Conference on Computer Communications Workshops (INFOCOM WKSHPS),
  IEEE, 2020, pp. 677--682.

\bibitem{keshavarz2018towards}
M.~Keshavarz, M.~Anwar, Towards improving privacy control for smart homes: A
  privacy decision framework, in: 2018 16th Annual Conference on Privacy,
  Security and Trust (PST), IEEE, 2018, pp. 1--3.

\bibitem{keshavarz2019automatic}
M.~Keshavarz, M.~Anwar, The automatic detection of sensitive data in smart
  homes, in: International Conference on Human-Computer Interaction, Springer,
  2019, pp. 404--416.

\bibitem{erdelj2017help}
M.~Erdelj, E.~Natalizio, K.~R. Chowdhury, I.~F. Akyildiz, {Help from the sky:
  Leveraging UAVs for disaster management}, IEEE Pervasive Computing 16~(1)
  (2017) 24--32.

\bibitem{Afghah_ACC}
F.~{Afghah}, M.~{Zaeri-Amirani}, A.~{Razi}, J.~{Chakareski}, E.~{Bentley}, A
  coalition formation approach to coordinated task allocation in heterogeneous
  uav networks, in: 2018 Annual American Control Conference (ACC), 2018, pp.
  5968--5975.
\newblock \href {http://dx.doi.org/10.23919/ACC.2018.8431278}
  {\path{doi:10.23919/ACC.2018.8431278}}.

\bibitem{aggarwal2020risk}
R.~Aggarwal, A.~Soderlund, M.~Kumar, D.~Grymin, Risk aware suas path planning
  in an unstructured wildfire environment, in: 2020 American Control Conference
  (ACC), IEEE, 2020, pp. 1767--1772.

\bibitem{Afghah_INFOCOM}
F.~{Afghah}, A.~{Razi}, J.~{Chakareski}, J.~{Ashdown}, Wildfire monitoring in
  remote areas using autonomous unmanned aerial vehicles, in: IEEE INFOCOM 2019
  - IEEE Conference on Computer Communications Workshops (INFOCOM WKSHPS),
  2019, pp. 835--840.
\newblock \href {http://dx.doi.org/10.1109/INFCOMW.2019.8845309}
  {\path{doi:10.1109/INFCOMW.2019.8845309}}.

\bibitem{afghah2018reputation}
F.~Afghah, A.~Shamsoshoara, L.~Njilla, C.~Kamhoua, A reputation-based
  stackelberg game model to enhance secrecy rate in spectrum leasing to selfish
  iot devices, in: IEEE INFOCOM 2018-IEEE Conference on Computer Communications
  Workshops (INFOCOM WKSHPS), IEEE, 2018, pp. 312--317.

\bibitem{shamsoshoara2019distributed}
A.~Shamsoshoara, M.~Khaledi, F.~Afghah, A.~Razi, J.~Ashdown, {Distributed
  cooperative spectrum sharing in uav networks using multi-agent reinforcement
  learning}, in: 2019 16th IEEE Annual Consumer Communications \& Networking
  Conference (CCNC), IEEE, 2019, pp. 1--6.

\bibitem{shamsoshoara2019solution}
A.~Shamsoshoara, M.~Khaledi, F.~Afghah, A.~Razi, J.~Ashdown, K.~Turck, {A
  solution for dynamic spectrum management in mission-critical UAV networks},
  in: 2019 16th Annual IEEE International Conference on Sensing, Communication,
  and Networking (SECON), IEEE, 2019, pp. 1--6.

\bibitem{shamsoshoara2020autonomous}
A.~Shamsoshoara, F.~Afghah, A.~Razi, S.~Mousavi, J.~Ashdown, K.~Turk, {An
  Autonomous Spectrum Management Scheme for Unmanned Aerial Vehicle Networks in
  Disaster Relief Operations}, IEEE Access 8 (2020) 58064--58079.

\bibitem{mousavi2019use}
S.~Mousavi, F.~Afghah, J.~D. Ashdown, K.~Turck, Use of a quantum genetic
  algorithm for coalition formation in large-scale uav networks, Ad Hoc
  Networks 87 (2019) 26--36.

\bibitem{mousavi2018leader}
S.~Mousavi, F.~Afghah, J.~D. Ashdown, K.~Turck, Leader-follower based coalition
  formation in large-scale uav networks, a quantum evolutionary approach, in:
  IEEE INFOCOM 2018-IEEE Conference on Computer Communications Workshops
  (INFOCOM WKSHPS), IEEE, 2018, pp. 882--887.

\bibitem{mousavi2017traffic}
S.~S. Mousavi, M.~Schukat, E.~Howley, Traffic light control using deep
  policy-gradient and value-function-based reinforcement learning, IET
  Intelligent Transport Systems 11~(7) (2017) 417--423.

\bibitem{mousavi2016deep}
S.~S. Mousavi, M.~Schukat, E.~Howley, Deep reinforcement learning: an overview,
  in: Proceedings of SAI Intelligent Systems Conference, Springer, 2016, pp.
  426--440.

\bibitem{sarcinelli2019handling}
R.~Sarcinelli, R.~Guidolini, V.~B. Cardoso, T.~M. Paix{\~a}o, R.~F. Berriel,
  P.~Azevedo, A.~F. De~Souza, C.~Badue, T.~Oliveira-Santos, Handling
  pedestrians in self-driving cars using image tracking and alternative path
  generation with fren{\'e}t frames, Computers \& Graphics 84 (2019) 173--184.

\bibitem{mousavi2016learning}
S.~Mousavi, M.~Schukat, E.~Howley, A.~Borji, N.~Mozayani, Learning to predict
  where to look in interactive environments using deep recurrent q-learning,
  arXiv preprint arXiv:1612.05753.

\bibitem{wang2019benchmarking}
Y.~E. Wang, G.-Y. Wei, D.~Brooks, {Benchmarking TPU, GPU, and CPU platforms for
  deep learning}, arXiv preprint arXiv:1907.10701.

\bibitem{google_edge_tpu:online}
{Edge TPU - Google}, \url{https://cloud.google.com/edge-tpu}, (Accessed on
  10/29/2020).

\bibitem{NVIDIAJe4:online}
{NVIDIA Jetson Nano},
  \url{https://developer.nvidia.com/embedded/jetson-nano-developer-kit},
  (Accessed on 10/29/2020).

\bibitem{wu2020transfer}
H.~Wu, H.~Li, A.~Shamsoshoara, A.~Razi, F.~Afghah, {Transfer Learning for
  Wildfire Identification in UAV Imagery}, in: 2020 54th Annual Conference on
  Information Sciences and Systems (CISS), IEEE, 2020, pp. 1--6.

\bibitem{Phantom345:phantom3}
{DJI - Phantom 3 Professional}, \url{https://www.dji.com/phantom-3-pro},
  (Accessed on 08/30/2020).

\bibitem{DJITheWo79:Matrice200}
D.~Company, {DJI - Matrice 200 V1},
  \url{https://www.dji.com/matrice-200-series/info\#specs}, (Accessed on
  08/30/2020).

\bibitem{DJIZenmu25:zenmuse}
{DJI Zenmuse X4S - Specifications, FAQs, Videos, Tutorials, Manuals, DJI GO -
  DJI}, \url{https://www.dji.com/zenmuse-x4s/info\#specs}, (Accessed on
  08/30/2020).

\bibitem{FLIRVueP5:online}
{FLIR Vue Pro R Radiometric Drone Thermal Camera | FLIR Systems},
  \url{https://www.flir.com/products/vue-pro-r/}, (Accessed on 12/08/2020).

\bibitem{qad6-r683-20}
A.~Shamsoshoara, F.~Afghah, A.~Razi, L.~Zheng, P.~Fulé, E.~Blasch,
  \href{https://dx.doi.org/10.21227/qad6-r683}{{Aerial images for Pile burn
  detection using drones (UAVs)}} (2020).
\newblock \href {http://dx.doi.org/10.21227/qad6-r683}
  {\path{doi:10.21227/qad6-r683}}.
\newline\urlprefix\url{https://dx.doi.org/10.21227/qad6-r683}

\bibitem{youtube2020_dataset}
A.~Shamsoshoara, {Aerial Images for Pile burn Detection Using Drones (UAVs)},
  \url{https://youtu.be/bHK6g37_KyA}, wireless Networking \& Information
  Processing (WINIP) LAB, accessed on 11/19/2020 (2020).

\bibitem{celik2009fire}
T.~Celik, H.~Demirel, {Fire detection in video sequences using a generic color
  model}, Fire safety journal 44~(2) (2009) 147--158.

\bibitem{umar2017state}
M.~M. Umar, L.~C.~D. Silva, M.~S.~A. Bakar, M.~I. Petra, {State of the art of
  smoke and fire detection using image processing}, International Journal of
  Signal and Imaging Systems Engineering 10~(1-2) (2017) 22--30.

\bibitem{binti2015fire}
N.~I. binti Zaidi, N.~A.~A. binti Lokman, M.~R. bin Daud, H.~Achmad, K.~A.
  Chia, {Fire recognition using RGB and YCbCr color space}, ARPN Journal of
  Engineering and Applied Sciences 10~(21) (2015) 9786--9790.

\bibitem{yuan2015uav}
C.~Yuan, Z.~Liu, Y.~Zhang, {UAV-based forest fire detection and tracking using
  image processing techniques}, in: 2015 International Conference on Unmanned
  Aircraft Systems (ICUAS), IEEE, 2015, pp. 639--643.

\bibitem{qi2009computer}
X.~Qi, J.~Ebert, {A computer vision based method for fire detection in color
  videos}, International journal of imaging 2~(S09) (2009) 22--34.

\bibitem{kundu2018highly}
S.~Kundu, V.~Mahor, R.~Gupta, {A highly accurate fire detection method using
  discriminate method}, in: 2018 International Conference on Advances in
  Computing, Communications and Informatics (ICACCI), IEEE, 2018, pp.
  1184--1189.

\bibitem{chollet2017xception}
F.~Chollet, {Xception: Deep learning with depthwise separable convolutions},
  in: Proceedings of the IEEE conference on computer vision and pattern
  recognition, 2017, pp. 1251--1258.

\bibitem{szegedy2016rethinking}
C.~Szegedy, V.~Vanhoucke, S.~Ioffe, J.~Shlens, Z.~Wojna, {Rethinking the
  inception architecture for computer vision}, in: Proceedings of the IEEE
  conference on computer vision and pattern recognition, 2016, pp. 2818--2826.

\bibitem{szegedy2016inception}
C.~Szegedy, S.~Ioffe, V.~Vanhoucke, A.~Alemi, {Inception-v4, inception-resnet
  and the impact of residual connections on learning}, arXiv preprint
  arXiv:1602.07261.

\bibitem{he2016deep}
K.~He, X.~Zhang, S.~Ren, J.~Sun, {Deep residual learning for image
  recognition}, in: Proceedings of the IEEE conference on computer vision and
  pattern recognition, 2016, pp. 770--778.

\bibitem{li2017convergence}
Y.~Li, Y.~Yuan, {Convergence analysis of two-layer neural networks with relu
  activation}, in: Advances in neural information processing systems, 2017, pp.
  597--607.

\bibitem{github:code_dataset_fire}
A.~Shamsoshoara,
  {Fire-Detection-UAV-Aerial-Image-Classification-Segmentation-UnmannedAerialVehicle},
  \url{https://github.com/AlirezaShamsoshoara/Fire-Detection-UAV-Aerial-Image-Classification-Segmentation-UnmannedAerialVehicle}
  (2020).

\bibitem{yuan2017aerial}
C.~Yuan, Z.~Liu, Y.~Zhang, {Aerial images-based forest fire detection for
  firefighting using optical remote sensing techniques and unmanned aerial
  vehicles}, Journal of Intelligent \& Robotic Systems 88~(2-4) (2017)
  635--654.

\bibitem{ccelik2007fire}
T.~{\c{C}}elik, H.~{\"O}zkaramanl{\i}, H.~Demirel, {Fire and smoke detection
  without sensors: Image processing based approach}, in: 2007 15th European
  Signal Processing Conference, IEEE, 2007, pp. 1794--1798.

\bibitem{khalil2020fire}
A.~Khalil, S.~U. Rahman, F.~Alam, I.~Ahmad, I.~Khalil, {Fire Detection Using
  Multi Color Space and Background Modeling}, Fire Technology (2020) 1--19.

\bibitem{Labelbox:online}
{Image segmentation software from Labelbox},
  \url{https://labelbox.com/product/image-segmentation}, (Accessed on
  10/25/2020).

\bibitem{Django:online}
{image labelling tool: Bitbucket},
  \url{https://bitbucket.org/ueacomputervision/image-labelling-tool/src/master/},
  (Accessed on 10/25/2020).

\bibitem{labelImg:online}
{labelImg: LabelImg is a graphical image annotation tool and label object
  bounding boxes in images}, \url{https://github.com/tzutalin/labelImg},
  (Accessed on 10/25/2020).

\bibitem{MATLABImageLabeler:online}
{Image Labeler, MATLAB \& Simulink},
  \url{https://www.mathworks.com/help/vision/ug/get-started-with-the-image-labeler.html},
  (Accessed on 10/25/2020).

\bibitem{GIMPGNUI35:online}
{GIMP - GNU Image Manipulation Program}, \url{https://www.gimp.org/}, (Accessed
  on 10/25/2020).

\bibitem{ronneberger2015u}
O.~Ronneberger, P.~Fischer, T.~Brox, {U-net: Convolutional networks for
  biomedical image segmentation}, in: International Conference on Medical image
  computing and computer-assisted intervention, Springer, 2015, pp. 234--241.

\bibitem{ELU:online}
{ELU Activation Function, {ML} Glossary documentation},
  \url{https://ml-cheatsheet.readthedocs.io/en/latest/activation\_functions.html\#elu},
  (Accessed on 10/26/2020).

\end{thebibliography}

\end{document}